\documentclass{llncs}

\usepackage{graphicx,epstopdf}
\graphicspath{{./pdf/}{../jpeg/}{../eps/}{./Figures/}}
\DeclareGraphicsExtensions{.pdf,.jpeg,.png,.eps}

\usepackage[cmex10]{amsmath}
\usepackage[caption=false,font=footnotesize]{subfig}
\usepackage[ruled,vlined]{algorithm2e}
\usepackage{url}
\usepackage{multirow}

\begin{document}

\title{Maximizing Diversity for Multimodal Optimization}

\author{Fabr\'{i}cio Olivetti de Fran\c{c}a}
\institute{CMCC, Federal University of ABC (UFABC) -- Brazil \\
 \email{folivetti@ufabc.edu.br}
}

\maketitle



\vspace{-5pt}
\section{Introduction}
\label{sec:introduction}
\vspace{-5pt}

Most multimodal optimization algorithms use the so called \textit{niching methods}~\cite{mahfoud1995niching} in order to promote diversity during optimization, while others, like \textit{Artificial Immune Systems}~\cite{de2010conceptual} try to find multiple solutions as its main objective. One of such algorithms, called \textit{dopt-aiNet}~\cite{de2005artificial}, introduced the Line Distance that measures the distance between two solutions regarding their basis of attraction. In this short abstract I propose the use of the Line Distance measure as the main objective-function in order to locate multiple optima at once in a population.

\vspace{-5pt}
\subsection{Line Distance}
\vspace{-5pt}

The Line Distance between two solutions $x$ and $y$ is calculated by taking the middle point $z = (x+y)/2$ and building three $(n+1)$-dimensional vectors $x' = [x, f(x)]$, the distance between the point $z'$ and the line formed by $x'$ and $y'$ is then calculated. If both $x$ and $y$ are at the same optimum, the line formed between them will be very close to the function contour resulting in a small distance, but if they are at different optima, it will be proportional to $f(x)$.

\vspace{-5pt}
\subsection{Maximizing Diversity}
\vspace{-5pt}

The maximization of the Line Distance between two points will have its maximum when both points are at different optima, so if we maximize this objective, it is expected that the algorithm returns several different optima. This objective-function may be used as the main goal of the optization process or as a supporting operator (i.e., crossover, mutation) for any populational algorithm.

To test this claim, we can use this simples algorithm: create an initial population $P$ with a single solution drawn uniformly at random on the problem domain. Next, for a given number of iterations repeat two procedures: expansion of current solutions and suppression of similar solutions.

During the expansion process, for each solution $x$, $m$ new solutions are created by optimizing $\underset{\alpha} {\mathrm{argmax}} ld(x + \alpha.d/\|\|d\|\|,x)$, where $d$ is a random direction drawn uniformly at random from $[-1,1]^n$, $\alpha$ is the step to walk into the given direction, that can be found by any unidimensional search method, and $ld(x,y)$ is the line distance between $x$ and $y$. Each new solution is aggregated into the  population if its distance to the original solution $x$ is greater than a threshold $\sigma$. If for a given solution, all of its generated solutions are discarded, then this solution is moved to the \textbf{local optima} population $LP$ as it is assumed that all the nearby optima around this solution was already discovered.

At the suppression step, all the solutions of each of the populations are compared to each other, whenever two solutions have distance greater than $\sigma$, the worst of the two is discarded. Notice that the optimization process is indirectly performed by the distance function that has its maximum whenever one of the solutions is located at a local optima.

\vspace{-5pt}
\section{Experimental Results}
\label{sec:experiments}
\vspace{-5pt}

\begin{figure}
\vspace{-25pt}
\subfloat[Rastrigin]{\includegraphics[trim=1cm 1cm 1cm 1cm, clip=true,width=0.4\textwidth]{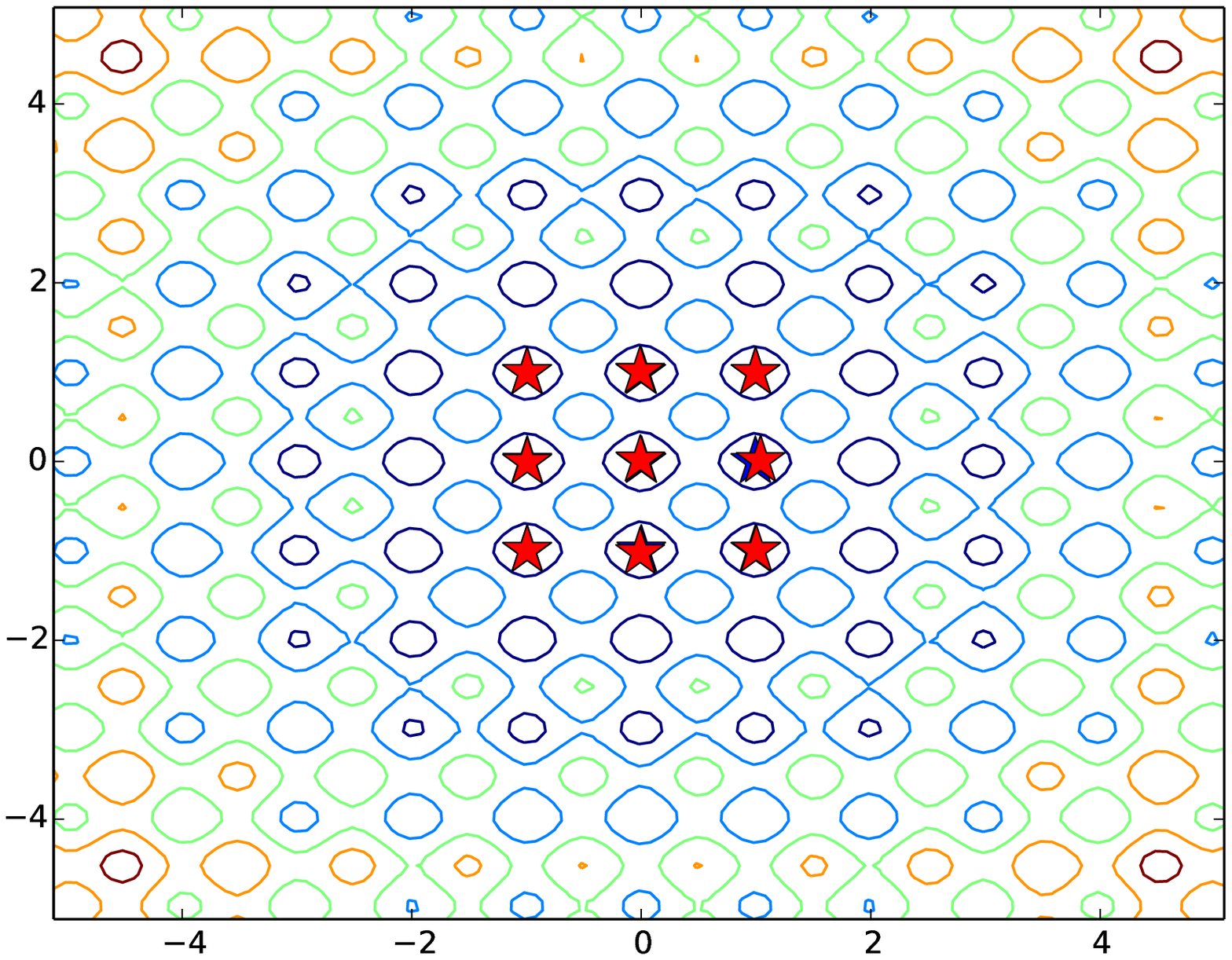}}
\subfloat[Grienwank]{\includegraphics[trim=1cm 1cm 1cm 1cm, clip=true,width=0.4\textwidth]{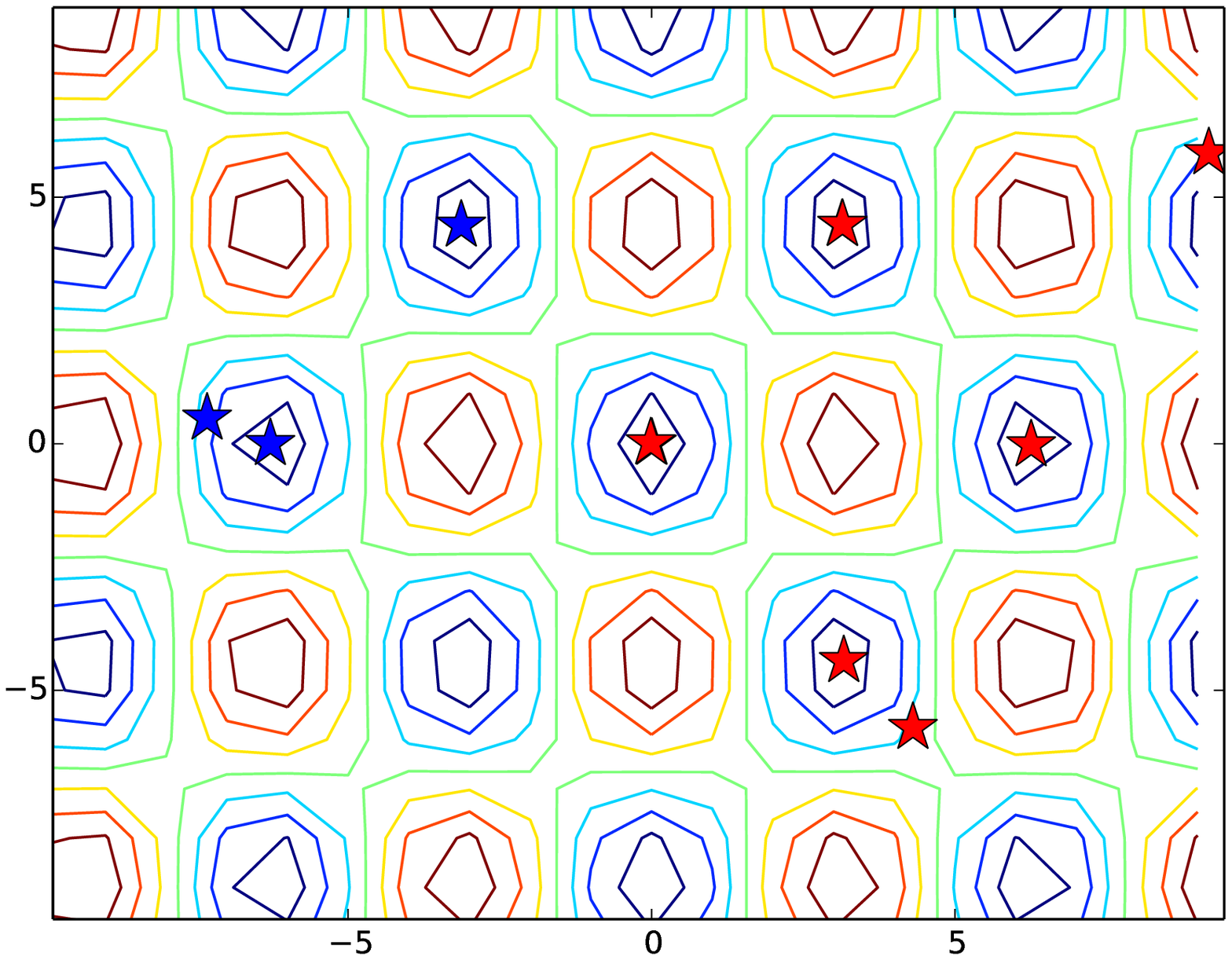}}
\caption{Some of the optima found by maximizing diversity.}
\label{fig:optima}
\vspace{-25pt}
\end{figure}

For a simple experiment it was chosen two well-known multimodal functions: Rastrigin and Grienwank, those functions were tested in $\Re^2$ and the parameters were $m=10$ for both function and, $\sigma=0.9$ for Rastrigin and $\sigma=0.1$ for Grienwank. For $10$ repetitions of $1,000$ iterations of this simple procedure could find $9$ different local optima for Rastrigin and $12$ for Griewank, in both cases the global optima was also found. The results are depicted in Fig.~\ref{fig:optima}.

\vspace{-5pt}
\section{Conclusion}
\label{sec:conclusion}
\vspace{-5pt}

In this extended abstract it is argued that the direct maximization of the diversity may be better suited for multimodal optimization algorithms. It is shown that by using the adequate diversity measure with a proper algorithm it is possible to find many different local optima. It must be noticed that some objective-functions may have an explosive growth on the number of local optima with the growth of dimension, so this procedure may be best suited as an evolutionary operator to promote diversity.
%
%

\bibliographystyle{splncs}
\bibliography{bigbang}

\end{document}